\documentclass[sigconf]{acmart}
\usepackage[flushleft]{threeparttable}

\AtBeginDocument{%
  \providecommand\BibTeX{{%
    \normalfont B\kern-0.5em{\scshape i\kern-0.25em b}\kern-0.8em\TeX}}}

\copyrightyear{2021}
\acmYear{2021}
\setcopyright{acmlicensed}\acmConference[CIKM '21]{Proceedings of the 30th ACM International Conference on Information and Knowledge Management}{November 1--5, 2021}{Virtual Event, QLD, Australia}
\acmBooktitle{Proceedings of the 30th ACM International Conference on Information and Knowledge Management (CIKM '21), November 1--5, 2021, Virtual Event, QLD, Australia}
\acmPrice{15.00}
\acmDOI{10.1145/3459637.3482152}
\acmISBN{978-1-4503-8446-9/21/11}

\begin{document}
\fancyhead{}
\title{Multivariate and Propagation Graph Attention Network for Spatial-Temporal Prediction with Outdoor Cellular Traffic}


\settopmatter{authorsperrow=1} 
\newcommand{\tsc}[1]{\textsuperscript{#1}} 

\author{Chung-Yi Lin\tsc{1,2}, Hung-Ting Su\tsc{1}, Shen-Lung Tung\tsc{2}, Winston H. Hsu\tsc{1,3}}
\affiliation{%
  \institution{\tsc{1}National Taiwan University,   \tsc{2}Chunghwa Telecom Laboratories,   \tsc{3}Mobile Drive Technology}
  \city{}
  \country{}
}
\renewcommand{\shortauthors}{Chung-Yi and Hung-Ting, et al.}

\begin{abstract}
 Spatial-temporal prediction is a critical problem for intelligent transportation, which is helpful for tasks such as traffic control and accident prevention. Previous studies rely on large-scale traffic data collected from sensors. However, it is unlikely to deploy sensors in all regions due to the device and maintenance costs. This paper addresses the problem via outdoor cellular traffic distilled from over two billion records per day in a telecom company, because outdoor cellular traffic induced by user mobility is highly related to transportation traffic. We study road intersections in urban and aim to predict future outdoor cellular traffic of all intersections given historic outdoor cellular traffic. Furthermore, we propose a new model for multivariate spatial-temporal prediction, mainly consisting of two extending graph attention networks (GAT). First GAT is used to explore correlations among multivariate cellular traffic. Another GAT leverages the attention mechanism into graph propagation to increase the efficiency of capturing spatial dependency. Experiments show that the proposed model significantly outperforms the state-of-the-art methods on our dataset.\footnote{The dataset is available on Github: https://github.com/cylin-cmlab/CIKM21-MPGAT}
\end{abstract}


\maketitle

\section{Introduction}
Recently, spatial-temporal prediction becomes one of the fundamental techniques in building intelligent transportation. Based on large-scale data (e.g., traffic speed, volume) collected from sensors, previous studies \cite{li2018diffusion,zheng2020gman} achieve successful performance. However, it is unlikely to deploy sensors in all regions (e.g., road intersections, rural areas) due to the device and maintenance costs, which gives rise to the task of data collection for spatial-temporal prediction. 

We seek an alternative approach to evaluate the traffic state without sensors. With billions of mobile devices entering the internet, massive records of cellular traffic \cite{fang2018mobile} are collected at cell towers. Many studies have been made for its application, such as cellular vehicle probes for traffic state estimation \cite{habtie2017artificial,valadkhani2017integration}, and cellular traffic prediction modeling \cite{wang2017spatiotemporal,fang2018mobile,wang2018spatio}. Nevertheless, existing studies rarely consider transportation traffic induced by user mobility. According to the analyses from over billions of records, outdoor cellular traffic is found to be induced by user mobility. Therefore, we leverage outdoor cellular traffic representing the traffic state.

In this paper, we propose a new spatial-temporal dataset via outdoor cellular traffic distilled from over a billion records per day in a telecom company, which is extremely valuable for surveilling traffic states in regions without sensors. For instance, if the accumulated outdoor cellular traffic in a unit time step exceeds the threshold, the region might occur certain events (e.g., traffic congestion or parade). We study the road intersections of a major city forming a road network and collect corresponding outdoor cellular traffic in time steps. This paper aims to predict the future outdoor cellular traffic of all intersections given historic outdoor cellular traffic.

Spatial-temporal prediction with a road network has been widely studied. Recent researches \cite{li2018diffusion,wu2019graph,wu2020connecting,tian2021spatial} have achieved great success in prediction by conducting propagating information on the graph data within graph neural networks (GNN). However, the dataset of previous studies is the traffic speed detected from sensors, which is usually less varied between time steps. While in our dataset, the quantity of outdoor cellular traffic can drastically vary in different times, especially the peak (e.g., 18:00) is 200 times greater than the least active times (e.g., 03:00) in the same road intersection, such significant difference is more challenging than that prior dataset. Therefore,  We argue that expanding the uni-historic data into multivariate consisting of various temporal-periodic data is critical to capturing more hidden correlations in the complex temporal pattern, which has not been explored in previous methods. As more complicated temporal features fed into the predictive model, we notice that the propagation process of GNN does not consider the dynamic attention between nodes during updating node information, which might reduce the capacity of modeling. 

To address the challenges, we propose a new framework for spatial-temporal prediction, namely \textbf{MPGAT} (Multivariate and Propagation Graph Attention Network), mainly consisting of two extending graph attention networks (GAT) \cite{velivckovic2018graph} modules. (1) Multivariate GAT (\textbf{M-GAT}) explores the correlations among multivariate input, which can be effectively adapted to multivariate time series. (2) Propagation GAT (\textbf{P-GAT}) incorporates the propagation strategy into GAT to captures the spatial dependency between regions, benefited from the attention mechanism and spatial closeness of the graph. We evaluate the proposed framework on our dataset. Experimental results with statistic analysis show that MPGAT outperforms significantly several state-of-the-art models.
\section{Dataset Description}
\textbf{Data Collection}:\ The data was collected from a large-scale cellular-geographic system in a telecom company. Each record contains International Mobile Station Equipment Identification (\textbf{IMEI}, a unique number of a mobile phone), the creation time, the GPS location, and the location type (categorized by the telecom company, e.g., outdoor and indoor). As privacy considerations, we use hashed IMEI to represent each mobile subscriber.
\\
\textbf{Spatial-temporal Dataset}:\ 
To explicitly discover the spatial dependency, we study six road intersections (e.g., around the train station and college) as a road network, as shown in Figure 1. Then, we individually aggregated the quantity of outdoor IMEI located at intersections in the unit time step (i.e., 5-minutes) as age spatial-temporal dataset. Specifically, the aggregated IMEI quantity over time steps demonstrates strong temporal correlations, and geographically connected intersections contain spatial dependence. 
\begin{figure}[h]
  \centering
  \includegraphics[width=0.8\linewidth]{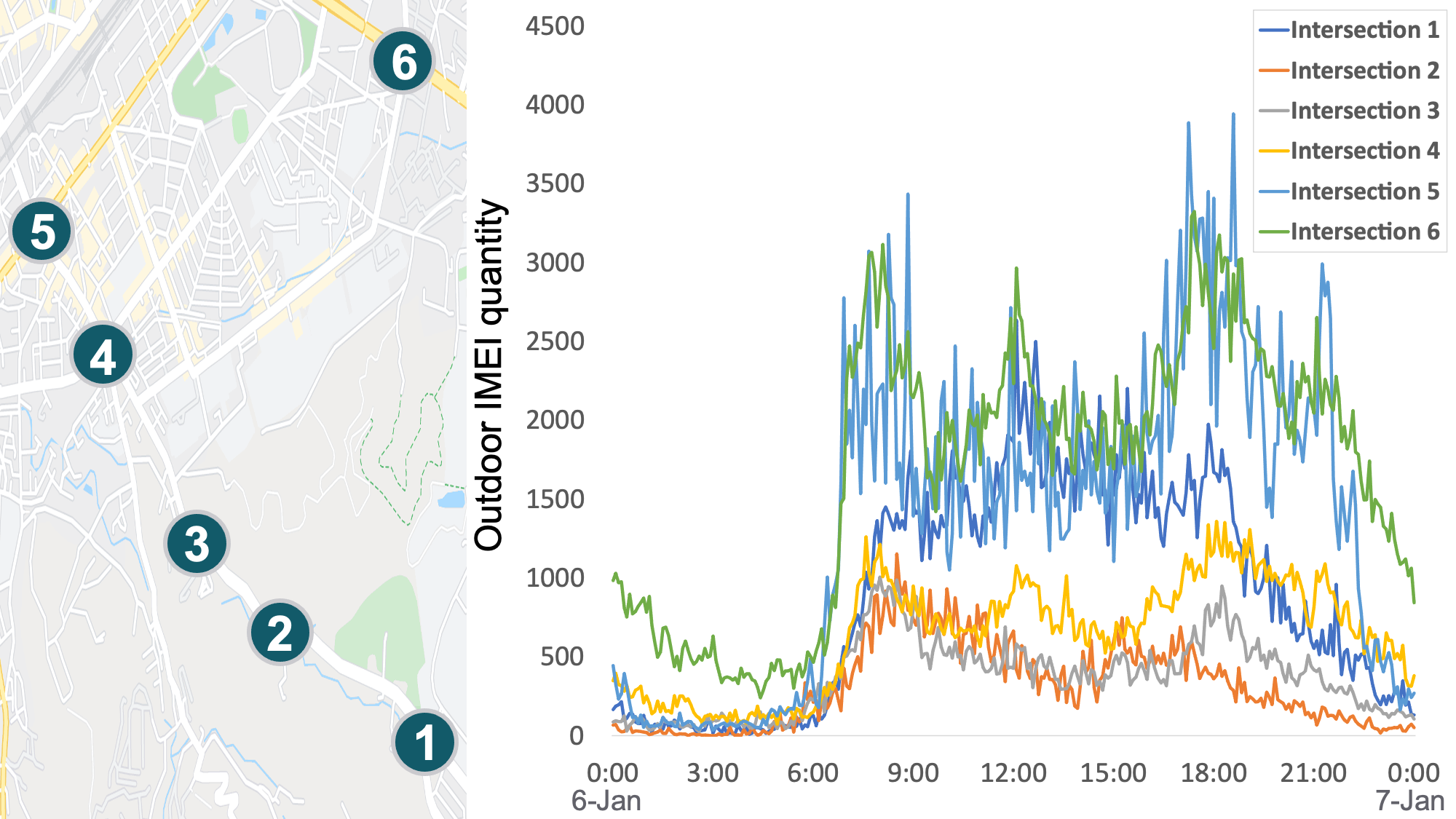}
  \caption{The dataset of outdoor cellular traffic is collected around the six road intersections(left fig.), and the IMEI quantities for every time step are shown in the right fig.}
  \Description{Description}
\end{figure}
\section{Preliminaries}
\textbf{Intersection Network}:\ Each road intersection is geographically connected. We define the network as a directed graph $G_{intersection}$ = ($V$,$E$), where $V$ is a set of $N$ intersections of the network, $E$ is the set of edges representing the connectivity between the intersections. 
\\
\textbf{Multivariate Input}:\ According to our dataset, IMEI quantity has a drastic change in different time steps. Following \cite{guo2019attention,lin2021predictions}, we expand IMEI quantity of unit time step as different temporal-periodic series to capture more temporal patterns, forming a multivariate input:
\\
\emph{(1)The IMEI quantity adjacent to predictions}:

$X_{q}=\{x_{q,t_{0}-T_{in}+1},...,x_{q,t_{0}}\}$$\in$$R^{N\times T_{in}}$ denotes historical IMEI quantity with $T_{in}$ time steps adjacent to the predictions, revealing the short-term factor, where $x_{q,t_{0}}$ is the IMEI quantity of $N$ intersections at time step $t_{0}$.
\\
\emph{(2)The moving average of IMEI quantity adjacent to predictions}:

$X_{ma_{T_{m}}}=\{x_{ma_{T_{m}},t_{0}-T_{in}+1},...,x_{ma_{T_{m}},t_{0}}\}$$\in$$R^{N\times T_{in}}$ is a set of moving average (MA) of IMEI quantity over $T_{in}$ time steps, where each value of $X_{ma_{T_{m}}}$ is calculated by accumulating the IMEI quantity over $T_{m}$ time steps and then dividing the sum by $T_{m}$. Moving average is commonly used with time-series data to smooth out short-term fluctuations and highlight longer-term trends \cite{yu2019taxi}. The features of $X_{ma_{5}}$  and $X_{ma_{20}}$ are adopted in this paper.
\\
\emph{(3)The daily intervals adjacent to prediction}: 

Suppose the sampling frequency is p time steps per day,
$X_{d}=\{x_{d,t_{0}-T_{in}\times p},x_{d,t_{0}-T_{in}-1\times p},...,x_{d,t_{0}}\}$$\in$$R^{N\times T_{in}}$  denotes a set of daily-interval data, consisting of IMEI quantity at $t_{0}$ time step in different days closed to predictions. For example, there are $p$(=288) time steps in one day by 5-minute time step. We consider the daily-interval quantity as one of the features to capture the sequence regular pattern.
\\
\textbf{Problem}:\ Given 
$X=\{X_{1},...,X_{F}\}$$\in$$R^{F\times N\times T_{in}}$ to denote the multivariate input of all the intersections over observed time steps, predict future IMEI quantity set
$Y=\{Y^{1},...,Y^{N}\}$$\in$$R^{N\times T_{out}}$  over the coming time steps, where $F$ is the number of features in multivariate input, $N$ is the number of intersections in $G_{intersection}$, $T_{in}$ and $T_{out}$ is the length of time steps, and  $Y^{k}$ is the future IMEI quantity of intersection $k$.
\section{Methodology}
Figure 2 illustrates the framework of our proposed Multivariate and Propagation Graph Attention Network (\textbf{MPGAT}). The multivariate input $X$, such as $X=\{X_{q},X_{ma_{5}},X_{ma_{20}},X_{d}\}$, is fed to M-GAT to capture correlations between multivariate input $X$. Then, the distilled outputs from M-GAT are forwarded to the connected  temporal convolution \cite{yu2016multi} (TCN) layer. For capturing the spatial-temporal dependency, TCN is interleaved with P-GAT as a spatial-temporal block, where P-GAT is to model the spatial dependency between intersections. By stacking multiple spatial-temporal blocks, the capacity of modeling the spatial-temporal dependencies increases \cite{wu2019graph}. To avoid gradient vanishing, Residual Layers \cite{He_2016_CVPR} is appended from the top of each interleaved layer to its end, and the Skipped Layers are concatenated after each TCN to an output layer.
\begin{figure}[h]
  \centering
  \includegraphics[width=1\linewidth]{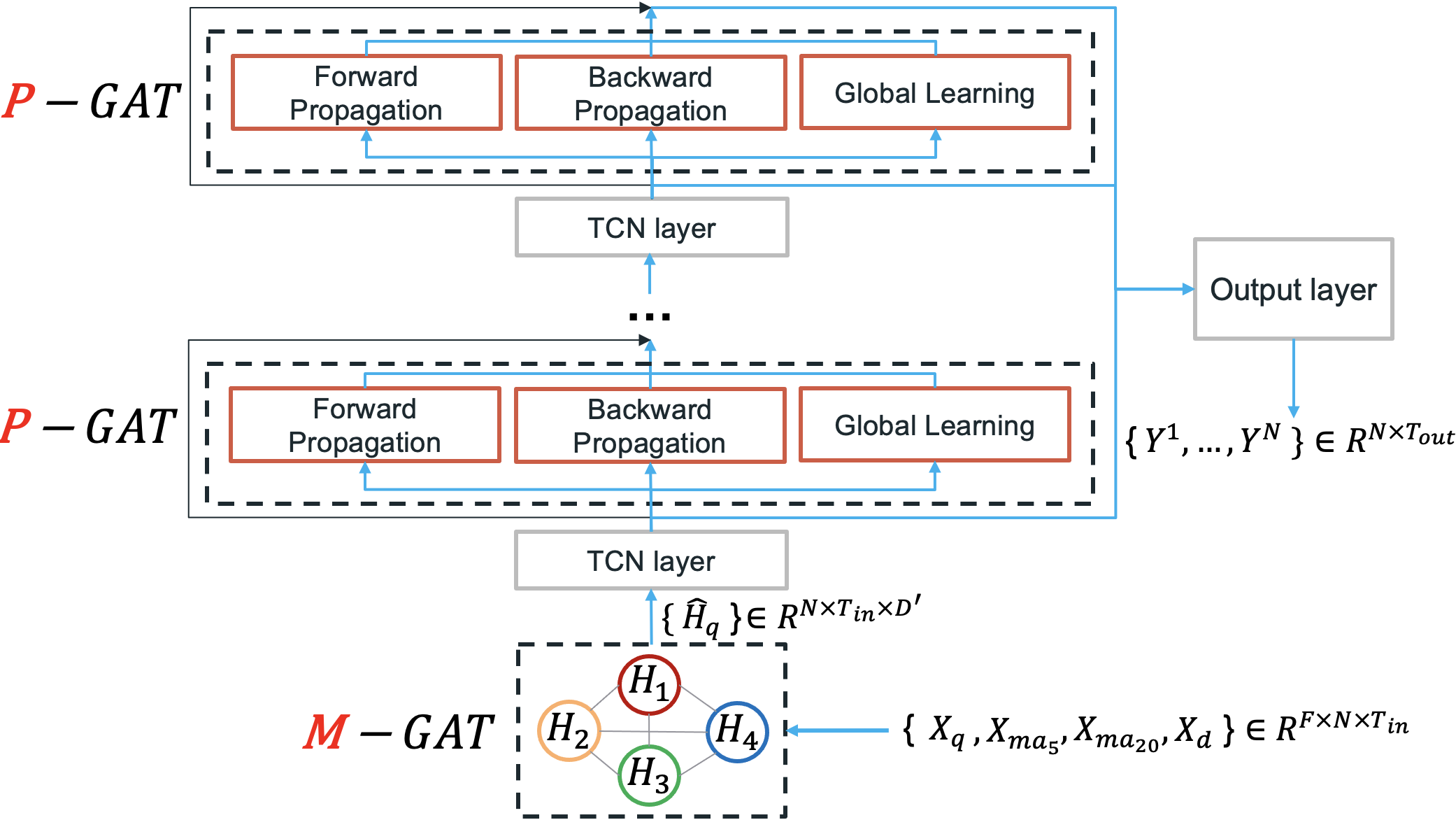}
  \caption{The framework of MPGAT, where M-GAT explores correlations among multivariate input, and P-GAT captures spatial dependency with multiple directions separately.}
  \Description{Description}
\end{figure}
\subsection{Multivariate Graph Attention (\textbf{M-GAT})}
As we aforementioned, the IMEI quantity series has been expanded into multivariate input. Due to the strong feature-extraction capability of GAT, M-GAT contains multiple GAT layers to capture the associations between multivariate input. Inspired by \cite{Huang_2019_ICCV,wang2019kgat}, M-GAT treats each component of multivariate input as one node in a complete graph and contains two stacked GAT layers. 

We consider a single layer of M-GAT as an example. For intersection $k$, the set of multivariate input 
$X^{k}=\{X_{1}^{\ \ k},...,X_{F}^{\ \ k}\}$ $\in$ $R^{F\times T_{in}}$ is first projected as a set of latent representations $Y^{k}=\{H_{1}^{\ \ k},...,H_{F}^{\ \ k}\}$ $\in$ $R^{F\times T_{in}\times D^{'}}$ by a convolution layer before M-GAT, where $T_{in}$ is the length of historical time steps, and $D^{'}$ is the dimension of latent representation. 
Afterward, $Y^{k}$ is fed to M-GAT, forming a complete graph with $F$ nodes. The attention score between nodes represents the importance from node $j$ to node $i$, where $i$,$j$ $\in\{1,2,…,F\}$, and can be computed as follows: 
\begin{equation}
  a_{ij}^{\ \ k}=\frac{exp(LeakyReLU(w_{c}^{T}(H_{i}^{\ \ k}\mathbin\Vert H_{j}^{\ \ k})))}{\sum_{j=1}^{F}exp(LeakyReLU(w_{c}^{T}(H_{i}^{\ \ k}\mathbin\Vert H_{j}^{\ \ k})))}
\end{equation}
, where $\mathbin\Vert$ is the concatenation operation, $w_{c}^{T}$$\in $$R^{2D^{'}}$ is the weight vector with transposition, and the attention score is normalized by a SoftMax function with LeakyReLU.

With the normalized attention scores, the output of one M-GAT layer for node $i$ is given by:
\begin{equation}
  \hat{H}_{i}^{\ \ k} =\sigma({\sum_{j=1}^{F}(a_{ij}^{\ \ k}H_{j}^{\ \ k})})
\end{equation}
, where $\sigma$ denotes a non-linear activation function. $\hat{H}_{i}^{\ \ k}$  is the aggregated latent representation for feature $i$ of intersection $k$, which contains the implicit influence from others. For $N$ intersections, the output set is presented as $\hat{H}=\{ \hat{H}_{1},...,\hat{H}_{F}  \}$ $\in$ $R^{F\times N\times T_{in}\times D^{'}}$. To reduce the model complexity, we only distill the latent representations of IMEI quantity outputted at the last layer of M-GAT, denotes as $\hat{H}_{q}$ $\in$ $R^{N\times T_{in}\times D^{'}}$, and feeding $\hat{H}_{q}$ into the next layer of MPGAT.
\begin{figure}[h]
  \centering
  \includegraphics[width=0.80\linewidth]{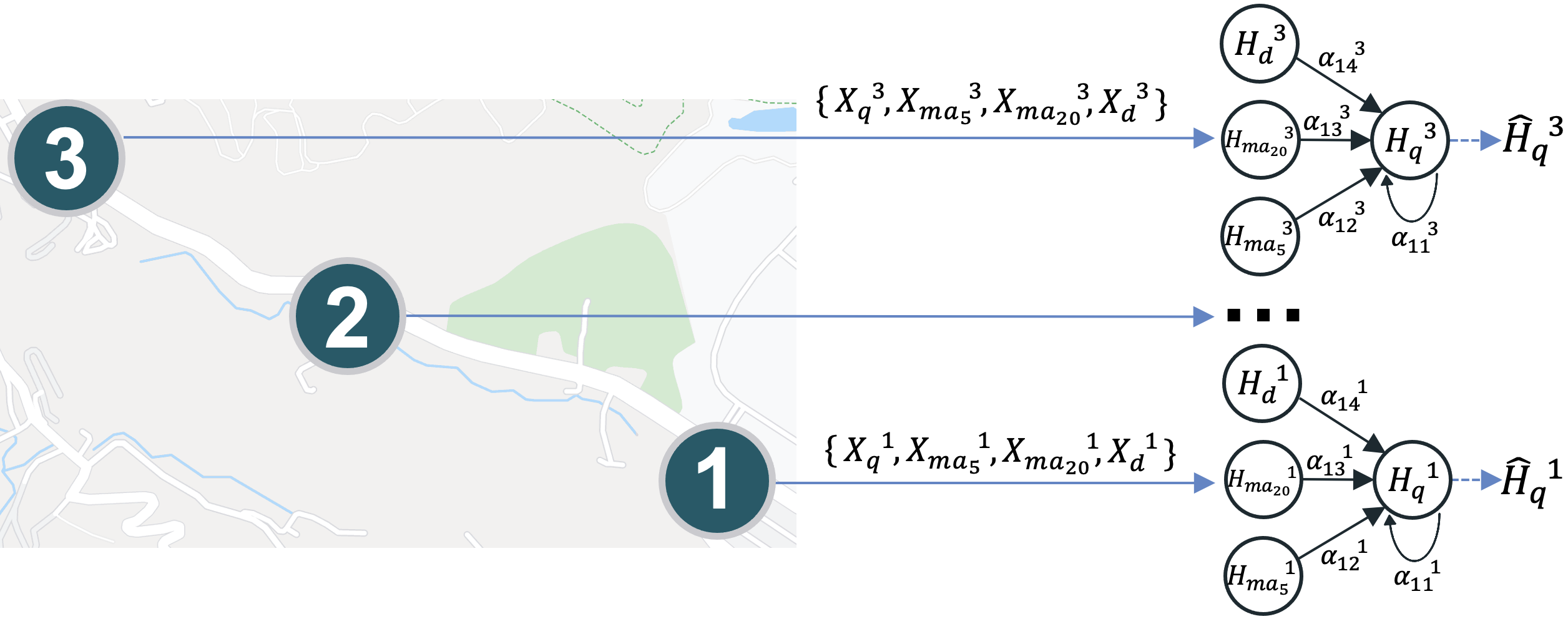}
  \caption{Multivariate input $X^k$ of each intersection $k$ is projected into latent representations $H^k$.}
  \Description{Description}
\end{figure}
\subsection{Propagation Graph Attention (\textbf{P-GAT})}
P-GAT consists of two directions \emph{Attention Propagation} and one \emph{Global learning} to capture the spatial dependency between intersections. To our best knowledge, we are the first work to apply the attention mechanism on graph propagation.
\\
\textbf{Attention Propagation}:\ The advantage of the propagation process \cite{wu2020connecting} is that it aggregates node information through the graph structure recursively and preserves a proportion of nodes’ states during the process, eliminating the smoothing problem with diffusion method in \cite{li2018diffusion}. With the benefit of propagating the states through graph structure, we generalize the attention mechanism into the propagation process, as shown in Figure 4. Considering $N$ nodes with the set of initial representations $V_{in}=\{V_{1},…,V_{N}\}$ from the previous layer, the propagation and output is defined as:
\begin{equation}
  V^{\mu}=(1-\beta)V_{in}+\beta(A_{mask}\ V^{\mu-1}) 
\end{equation}
\begin{equation}
  \hat{V}=\Delta(\mathbin\Vert_{\mu=0}^{U}V^{\mu})
\end{equation}
, where $V^{0}=V_{in}$, $A_{mask}$ is the \emph{masked attention matrix} for uni-direction, $\mu$ represents the propagation steps, $\beta$ is a hyperparameter between 0 and 1 to control the ratio of node’s propagation, $\Delta$ is a multilayer perceptron layer (MLP) to transform the channel dimension of concatenation results,  and $\hat{V}$ is the final output of $N$ nodes. In practice,  we employ the CNN as MLP, and set $U$ is 2. 

In case of the directed graph $G_{Intersection}$, P-GAT models two directions, the \emph{forward} and the \emph{backward} propagations with own masked attention matrix. Consider each intersection of $G_{Intersection}$ as one node, the attention score $a_{ij}$ in  uni-direction $A_{mask}$ between nodes $i$ and $j$, where $i$,$j$ $\in \{1,2,…,N\}$, is computed as:
\begin{table*}
  \caption{The average performance of MAPE comparison on outdoor cellular traffic dataset}
  \label{tab:commands}
  \begin{tabular}{p{2cm} p{2cm} p{1.2cm} p{2cm} p{1.2cm} p{2cm} p{1.2cm} p{2cm} p{1.2cm}  }
    \toprule 
    Models & Mean \textpm std. Dev. & h\tsc{1} & Mean \textpm std. Dev. & h\tsc{1} & Mean \textpm std. Dev. & h\tsc{1} & Mean \textpm std. Dev. & h\tsc{1} \\
    \midrule
     & 5min &  & 15min &  & 30min &  & 60min &  \\
    \midrule
    \texttt LSTM\cite{hochreiter1997long} & 0.1592±0.0034 & 0/\textbf{1} & 0.1801±0.0034	 & \textbf{1}/\textbf{1} & 0.1984±.0052 & \textbf{1}/\textbf{1} & 0.2342±0.0078 & \textbf{1}/\textbf{1}
    \\
    \texttt GWNET('19)\cite{wu2019graph} & 0.1576±0.0027 & -1/\textbf{1} & 0.1781±0.0034	 & 0/\textbf{1} & 0.1934±0.0028 & 0/\textbf{1} & 0.2194±0.0037 & 0/\textbf{1}
    \\ 
    \texttt MTGNN('20)\cite{wu2020connecting} & 0.1629±0.0032 & \textbf{1}/\textbf{1} & 0.1811±0.0027 & \textbf{1}/\textbf{1} & 0.1951±0.0031 & 0/\textbf{1} & 0.2198±0.0033 & 0/\textbf{1}
    \\
    \texttt STAWNET('21)\cite{tian2021spatial} & 0.1611±0.0026 &\textbf{1}/\textbf{1} &0.1823±0.0037 &\textbf{1}/\textbf{1} & 0.1947±0.0029 &0/\textbf{1} &0.2180±0.0026 &-1/\textbf{1}
    \\
    \texttt MPGAT-1 & 0.1582±0.0024 &  & 0.1778±0.0019 &  & 0.1940±0.0025 &  & 0.2213±0.0029	
    \\
    \texttt MPGAT & 0.1511±0.0013 &  & 0.1720±0.0012 &  & 0.1876±0.0016 &  & 0.2149±0.0026
    \\
    \bottomrule
  \end{tabular}
  
  \begin{tablenotes}
    \item h\tsc{1} means whether the result of MPGAT-1/MPGAT is significant according to Wilcoxon rank-sum test compared to the baseline method.
  \end{tablenotes}
\end{table*}
\begin{equation}
  e_{ij}=w_{p}^{T}(V_{i} \mathbin\Vert V_{j} ) 
\end{equation}
\begin{equation}
  a_{ij}=\frac{exp(LeakyReLU(e_{ij}))}{\sum_{j=1}^{N_{i}}exp(LeakyReLU(e_{ij}))}
\end{equation}
, where $V_{i}$ is the latent representation of node $i$ from the previous layer, $w_{p}^{T}\in R^{2D^{''}}$ is a weight vector with transposition, $D^{''}$ is the dimension of $V_{i}$, and $N_{i}$ denotes the adjacent neighbors of node $i$. The attention coefficient $e_{ij}$ in Equation 5 is masked according to the adjacency matrices in the corresponding direction, where $e_{ij}$ is masked with a large negative value (e.g., -9e15) if node $i$ and $j$ are not adjacent; otherwise, the coefficient $e_{ij}$ would be preserved. In this way, the attention score $a_{ij}$ between adjacent nodes would increase after SoftMax normalization, while the non-adjacent is set to zero. The matrix $A_{mask}$ can be represented as [$a_{ij}$], which implies the relationship between adjacent nodes but excludes the non-adjacent.
\begin{figure}[h]
  \centering
  \includegraphics[width=0.75\linewidth]{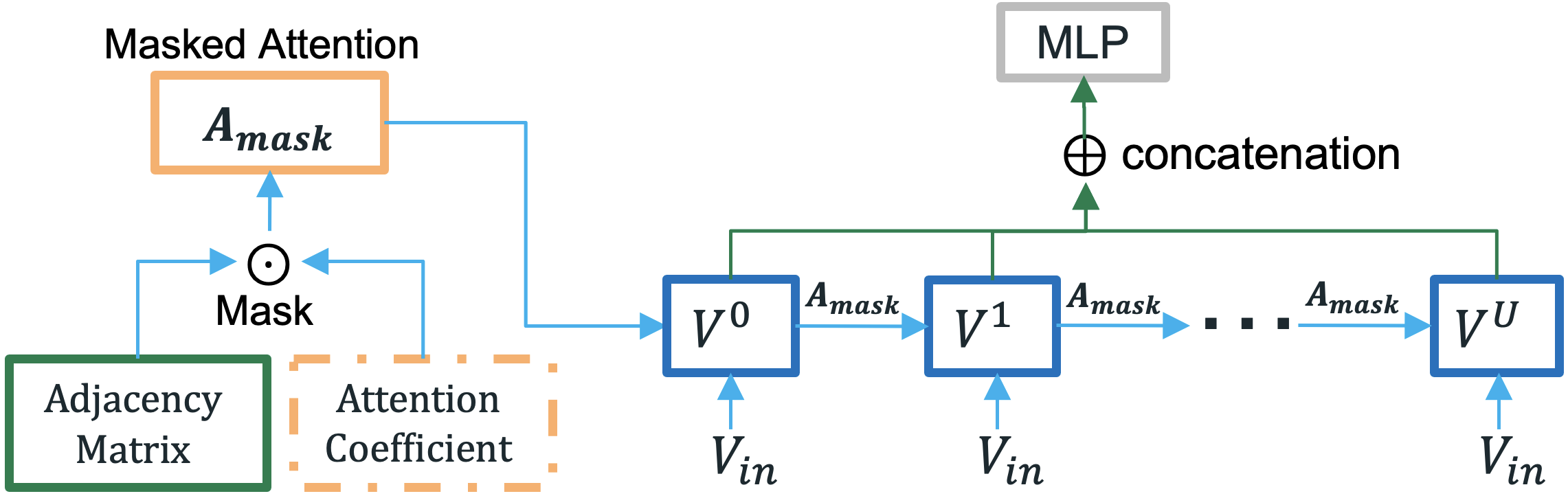}
  \caption{The function of attention propagation in P-GAT.}
  \Description{Description}
\end{figure}
\\
\textbf{Global Learning}:\ To discover hidden correlations among nodes, Global Learning treats each intersection as a node and builds a complete graph to capture the hidden relationships of spatial dependency. We use Equation 6 to construct an attention matrix, where $e_{ij}$ is not masked, $N_{i}$ denotes all the nodes of $G_{Intersection}$. All nodes would learn globally the attention score toward others and updates the node information directly.
\\
\textbf{Output}:\ P-GAT explores the bi-directional correlation with directional modeling and captures the implicit associations adaptively with global modeling for spatial dependency learning. Finally, P-GAT fuses two propagation modeling and global modeling outputs as an input fed to the next layer.

\section{Experiments}

\textbf{Dataset}:\ This paper verifies the proposed MPGAT and the compared models with the outdoor cellular traffic dataset, which contains the IMEI quantity in 5-minutes time steps of six road intersections ranging from Jan.1, 2020, to Jun.30, 2020. The dataset is split with 70\% for training, 10\% for validation, and 20\% for testing.
\\
\textbf{Comparison Methods}:\ Our dataset is a spatial-temporal dataset with a road network, similar tasks such as traffic speed have generally been addressed better by graph-based models \cite{zheng2020gman}. Due to the page limitation, we focus on the latest graph-based models as baselines in our evaluation: LSTM \cite{hochreiter1997long}, GWENT \cite{wu2019graph}, MTGNN \cite{wu2020connecting}, STAWNET \cite{tian2021spatial}. To verify the effectiveness of the correlation among multivariate, we build two models, MPGAT-1 and MPGAT, where  MPGAT-1 only adopt univariate IMEI quantity as input.
\\
\textbf{Experimental Settings}:\ The length of historical input $T_{in}$ is 12 for all models including ours and baselines. MPGAT uses eight spatial-temporal blocks to cover the input sequence, where each block has a TCN interleaved with a P-GAT. We adapt the Adam optimizer with a learning rate of 0.001 to train our model. The evaluation metrics we choose mean absolute percentage error (MAPE).
\\
\textbf{Main Results}:\ We implement our model and compare it with the baselines for  5 min,15 min, 30 min, and 60 min predictions on our dataset. We run 30 testing times and report the average values, as shown in Table 1. To determine whether MPGAT-1 and the MPGAT outperform in statistical significance with the compared models, we conducted the Wilcoxon rank-sum test inspired by \cite{juang2011adaptive}, where h = 1 is the proposed model significantly outperformed the compared model, h = -1 is the compared model significantly outperformed the proposed model, and h = 0 is not significantly different.

According to Table 1, as the prediction time increases, MAPE of the same method rises for all, showing that the longer the prediction time, the more challenging the prediction task. Second, with the statistical analysis, MPGAT significantly achieves the best performance in the dataset. Third, MPGAT-1 is slightly underperformed than GWNET while outperforming others in the short-term prediction. It is more beneficial to practical applications on short-term factors, e.g., transportation agencies can immediately optimize traffic congestion.

Figure 5 shows the changes in the prediction performance of various methods as the prediction steps increase. We observe that MPGAT consistently outperforms MPGAT-1 and baseline models, indicating it is critical to explore the correlation among multivariate, especially that our dataset has a drastic change between neighboring time steps. Moreover, we notice that the MAPE value of the same method conducted on our dataset is two times larger than the traffic speed dataset \cite{li2018diffusion}, which indicates that spatial-temporal prediction with cellular traffic data is more challenging. We are optimistic that our proposed task and dataset would pave a new path for spatial-temporal prediction and urban computing applications.
\begin{figure}[h]
  \centering
  \includegraphics[width=0.85\linewidth]{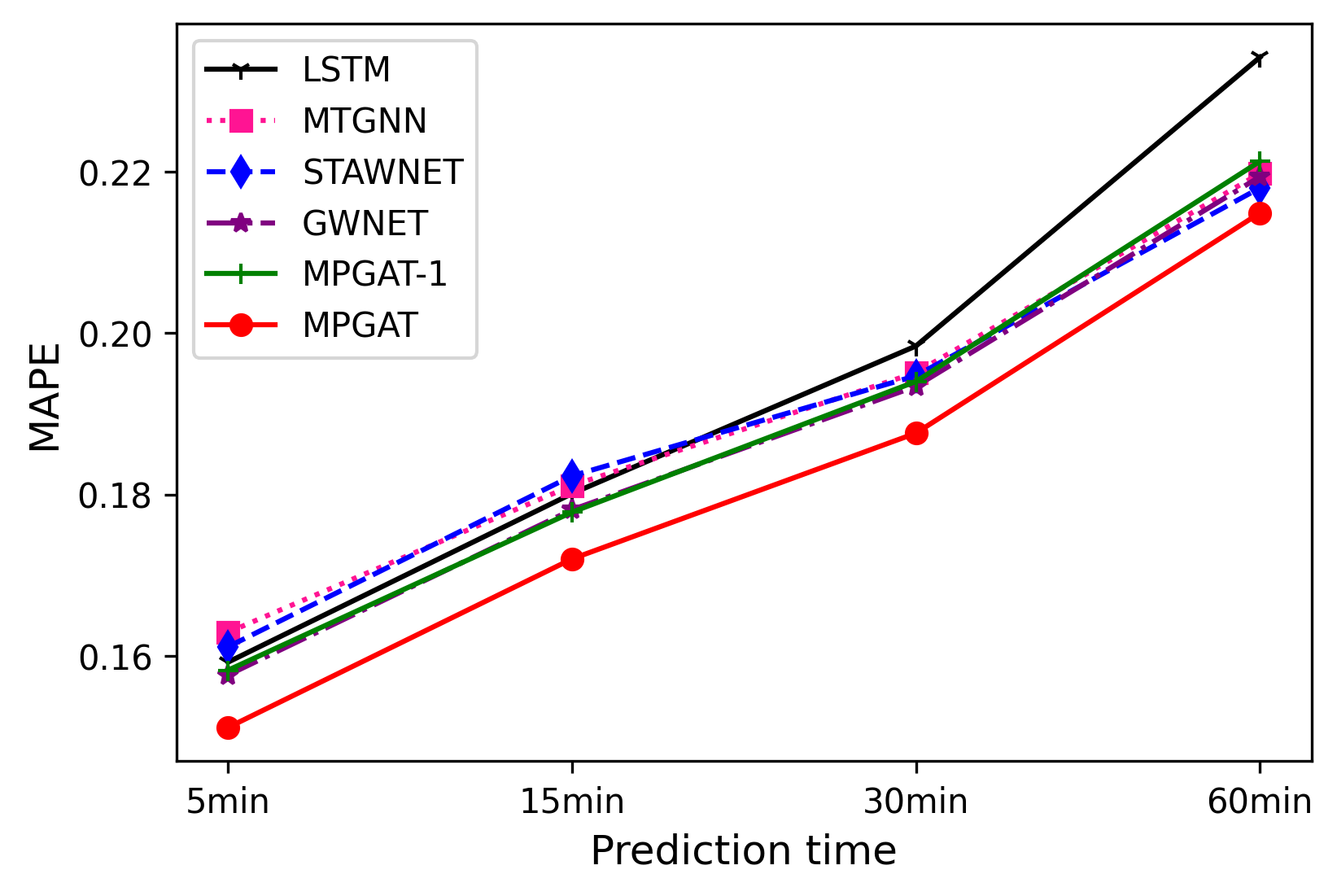}
  \caption{MAPE of compared methods in different prediction times shows MPGAT achieves better results in all cases.}
  \Description{Description}
\end{figure}
\section{Conclusion}
In this paper, we propose a new spatial-temporal dataset via outdoor cellular traffic and a model MPGAT for multivariate spatial-temporal prediction. Experimental results show that the proposed MPGAT significantly outperforms other models on the dataset.

\begin{acks}
This work was supported in part by the Ministry of Science and Technology, Taiwan, under Grant MOST 110-2634-F-002-026. We benefit from NVIDIA DGX-1 AI Supercomputer and are grateful to the National Center for High-performance Computing.
\end{acks}

\bibliographystyle{ACM-Reference-Format}
\bibliography{main}
\end{document}